\documentclass[10pt,twocolumn,letterpaper]{article}

\usepackage{cvpr}
\usepackage{times}
\usepackage{epsfig}
\usepackage{graphicx}
\graphicspath{{./images/}}
\usepackage{amsmath}
\usepackage{amssymb}
\usepackage[caption=false,font=footnotesize]{subfig}
\usepackage{booktabs}
\usepackage{xcolor}
\usepackage{url}

\usepackage[pagebackref=true,breaklinks=true,letterpaper=true,colorlinks,bookmarks=false]{hyperref}
\usepackage[capitalise]{cleveref}

\cvprfinalcopy 

\def\cvprPaperID{9831} 

\ifcvprfinal\fi
\begin{document}
	
	\title{FastDVDnet: Towards Real-Time Deep Video Denoising Without Flow Estimation}
	
	\author{Matias~Tassano\\
		GoPro France\\
		{\tt\small mtassano@gopro.com}
		\and
		Julie Delon\\
		MAP5, Université de Paris \& IUF\\
		{\tt\small julie.delon@parisdescartes.fr}
		\and
		Thomas Veit\\
		GoPro France\\
		{\tt\small tveit@gopro.com}
	}
	
	\maketitle
	
	\begin{abstract}
		In this paper, we propose a state-of-the-art video denoising algorithm based on a convolutional neural network architecture. Until recently, video denoising with neural networks had been a largely under explored domain, and existing methods could not compete with the performance of the best patch-based methods. The approach we introduce in this paper, called FastDVDnet, shows similar or better performance than other state-of-the-art competitors with significantly lower computing times. In contrast to other existing neural network denoisers, our algorithm exhibits several desirable properties such as fast runtimes, and the ability to handle a wide range of noise levels with a single network model. The characteristics of its architecture make it possible to avoid using a costly motion compensation stage while achieving excellent performance. The combination between its denoising performance and lower computational load makes this algorithm attractive for practical denoising applications. We compare our method with different state-of-art algorithms, both visually and with respect to objective quality metrics.
		
	\end{abstract}
	
	\section{Introduction}
	\label{sec:intro}
	Despite the immense progress made in recent years in  photographic sensors, noise reduction remains an essential step in video processing, especially when shooting conditions are challenging (low light, small sensors, etc.).
	
	Although image denoising has remained a very active research field through the years, too little work has been devoted to the restoration of digital videos. It should be noted, however, that some crucial aspects differentiate these two problems. On the one hand, a video contains much more information than a still image, which could help in the restoration process. On the other hand, video restoration requires good temporal coherency, which makes the restoration process much more demanding. Furthermore, since all recent cameras produce videos in high definition---or even larger---very fast and efficient algorithms are needed.
	
	
	In this paper we introduce another network for deep video denoising: FastDVDnet. This algorithm builds on DVDnet~\cite{Tassano2019}, but at the same time introduces a number of important changes with respect to its predecessor. Most notably, instead of employing an explicit motion estimation stage, the algorithm is able to implicitly handle motion thanks to the traits of its architecture. This results in a state-of-the-art algorithm which outputs high quality denoised videos while featuring very fast running times---even thousands of times faster than other relevant methods.
	
	\subsection{Image denoising}
	\label{sec:image-denoising}
	Contrary to video denoising, image denoising has enjoyed consistent popularity in past years. A myriad of new image denoising methods based on deep learning techniques have drawn considerable attention due to their outstanding performance. Schmidt and Roth proposed in~\cite{Schmidt2014a} the cascade of shrinkage fields method. The trainable nonlinear reaction diffusion model proposed by Chen and Pock in~\cite{Chen2017} builds on the former. In~\cite{Burger2012}, a multi-layer perceptron was successfully applied for image denoising.
	Methods such as these achieve performances comparable to those of well-known patch-based algorithms such as BM3D~\cite{Dabov2007a} or non-local Bayes (NLB~\cite{Lebrun2013c}). However, their limitations include performance restricted to specific forms of prior, or the fact that a different set of weights must be trained for each noise level.
	
	Another widespread approach involves the use of convolutional neural networks (CNN), e.g.\ RBDN~\cite{Santhanam2016}, MWCNN~\cite{Liu2018}, DnCNN~\cite{Zhang2017}, and FFDNet~\cite{Zhang2017a}. Their performance compares favorably to other state-of-the-art image denoising algorithms, both quantitatively and visually. These methods are composed of a succession of convolutional layers with nonlinear activation functions in between them.
	%
	%
	A salient feature that these CNN-based methods present is the ability to denoise several levels of noise with only one trained model. Proposed by Zhang \etal in~\cite{Zhang2017}, DnCNN is an end-to-end trainable deep CNN for image denoising. One of its main features is that it implements residual learning~\cite{He2016}, i.e.\ it estimates the noise existent in the input image rather than the denoised image. In a following paper~\cite{Zhang2017a}, Zhang \etal proposed FFDNet, which builds upon the work done for DnCNN\@.
	%
	More recently, the approaches proposed in~\cite{Plotz2018, Liu2018non} combine neural architectures with non-local techniques.
	
	\subsection{Video denoising}
	\label{sec:video-denoising}
	Video denoising is much less explored in the literature. The majority of recent video denoising methods are patch-based. We note in particular an extension of the popular BM3D to video denoising, V-BM4D~\cite{Maggioni2012}, and Video non-local Bayes (VNLB~\cite{Arias2018}). Neural network methods for video denoising have been even rarer than patch-based approaches. The algorithm in~\cite{chen2016deep} by Chen \etal is one of the first to approach this problem with recurrent neural networks. However, their algorithm only works on grayscale images and it does not achieve satisfactory results, probably due to the difficulties associated with training recurring neural networks~\cite{pascanu2013difficulty}. Vogels \etal proposed in~\cite{vogels2018denoising} an architecture based on kernel-predicting neural networks able to denoise Monte Carlo rendered sequences. The Video Non-Local Network (VNLnet~\cite{Davy2019}) fuses a CNN with a self-similarity search strategy. For each patch, the network finds the most similar patches via its first non-trainable layer, and this information is later used by the CNN to predict the clean image. In~\cite{Tassano2019}, Tassano \etal proposed DVDnet, which splits the denoising of a given frame in two separate denoising stages. Like several other methods, it relies on the estimation of motion of neighboring frames. Other very recent blind denoising approaches include the work by Ehret \etal~\cite{Ehret2019} and ViDeNN~\cite{Claus2019}. The latter shares with DVDnet the idea of performing denoising in two steps. However, contrary to DVDnet, ViDeNN does not employ motion estimation. Similarly to both DVDnet and ViDeNN, the use of spatio-temporal CNN blocks in restoration tasks has been also featured in~\cite{vogels2018denoising,Caballero2017}. Nowadays, the state-of-the-art is defined by DVDnet, VNLnet and VNLB. VNLB and VNLnet show the best performances for small values of noise, while DVDnet yields better results for larger values of noise. Both DVDnet and VNLnet feature significantly faster inference times than VNLB. As we will see, the performance of the method we introduce in this paper compares to the performance of the state-of-the-art, while featuring even faster runtimes.
	
	\section{FastDVDnet}
	\label{sec:method}
	\begin{figure*}[!t]
		\centering
		\subfloat[]{\includegraphics[width=0.60\linewidth]{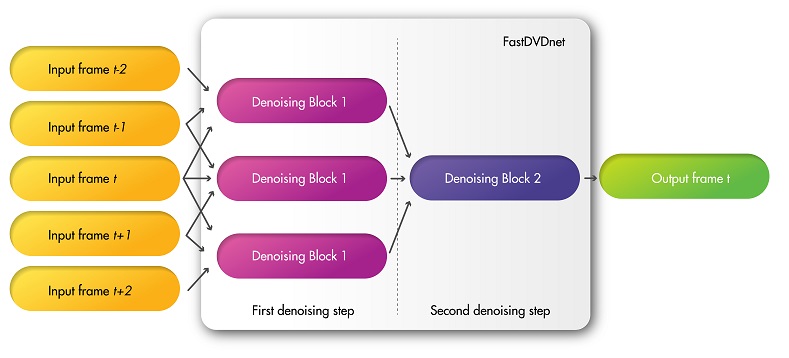}%
			\label{fig:diagram-overview}}
		\hfill
		\subfloat[]{\includegraphics[width=0.68\linewidth]{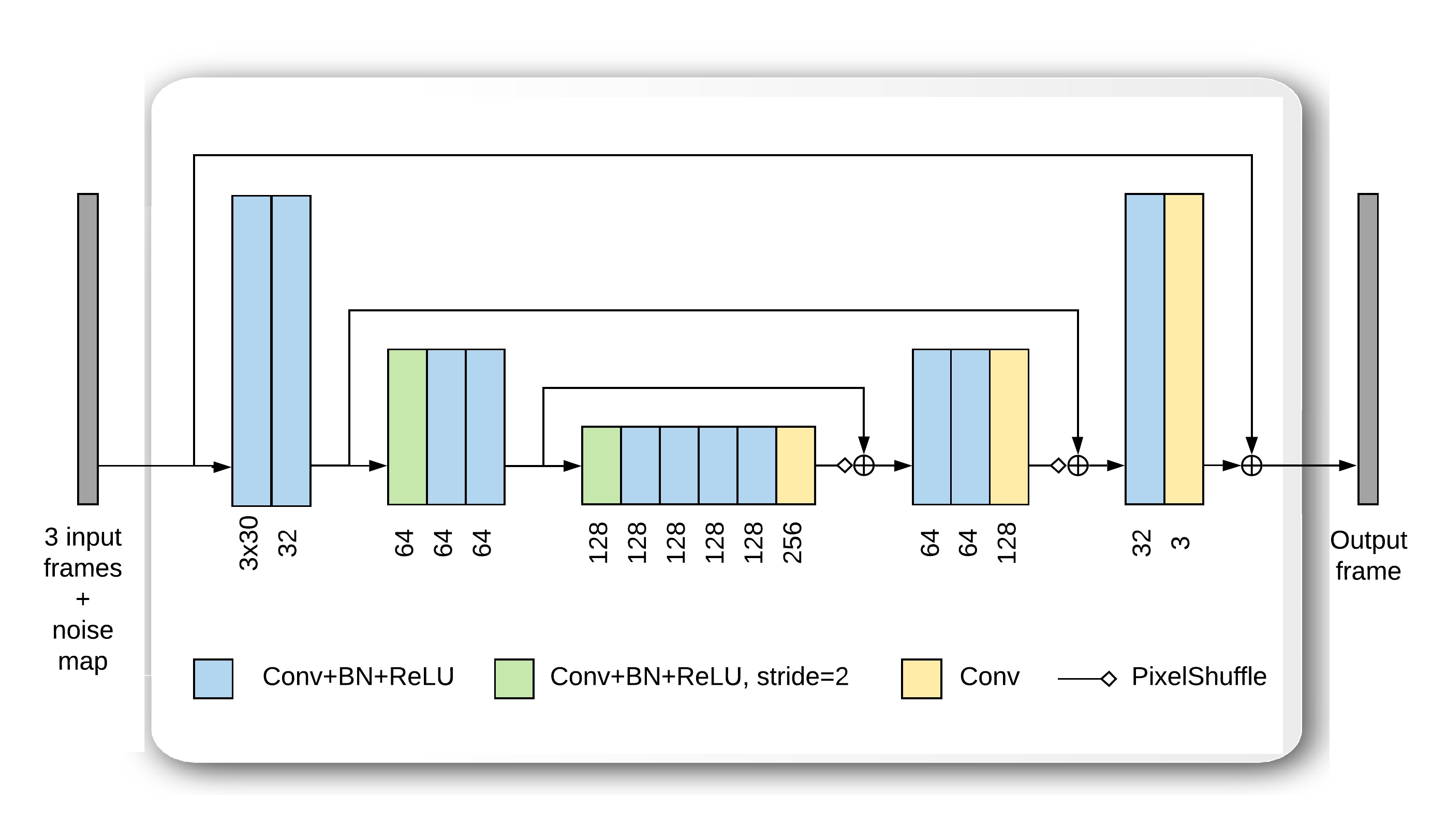}%
			\label{fig:diagram-denoiser}}
		
		\caption{\textit{Architecture used in FastDVDnet}. (a) A high-level diagram of the architecture. Five consecutive frames are used to denoise the middle frame. The frames are taken as triplets of consecutive frames and input to the \textit{Denoising Blocks 1}. The instances of these blocks have all the same weights. The triplet composed by the outputs of these blocks are used as inputs for \textit{Denoising Block 2}. The output of the latter is the estimate of the central input frame (\textit{Input frame t}). Both \textit{Denoising Block 1} and \textit{Denoising Block 2} share the same architecture, which is shown in (b). The denoising blocks of FastDVDnet are composed of a modified multi-scale U-Net.}
		\label{fig:architecture}
	\end{figure*}
	For video denoising algorithms, temporal coherence and flickering removal are crucial aspects in the perceived quality of the results~\cite{Seybold2018,Seshadrinathan2010}. In order to achieve these, an algorithm must make use of the temporal information existent in neighboring frames when denoising a given frame of an image sequence. In general, most previous approaches based on deep learning have failed to employ this temporal information effectively. Successful state-of-the-art algorithms rely mainly on two factors to enforce temporal coherence in the results, namely the extension of search regions from spatial neighborhoods to volumetric neighborhoods, and the use of motion estimation. 
	
	The use of volumetric (i.e.\ spatio-temporal) neighborhoods implies that when denoising a given pixel (or patch), the algorithm is going to look for similar pixels (patches) not only in the reference frame, but also in adjacent frames of the sequence. The benefits of this are two-fold. First, the temporal neighbors provide additional information which can be used to denoise the reference frame. Second, using temporal neighbors helps to reduce flickering as the residual error in each frame will be correlated.
	
	Videos feature a strong temporal redundancy along motion trajectories. This fact should facilitate denoising videos with respect to denoising images. Yet, this added information in the temporal dimension also creates an extra degree of complexity which could be difficult to tackle. In this context, motion estimation and/or compensation has been employed in a number of video denoising algorithms to help to improve denoising performance and temporal consistency~\cite{Liu2015,Tassano2019,Arias2018,Maggioni2012,Buades2016a}.
	
	We thus incorporated these two elements into our architecture. However, our algorithm does not include an explicit motion estimation/compensation stage. The capacity of handling the motion of objects is inherently embedded into the proposed architecture. Indeed, our architecture is composed of a number of modified U-Net~\cite{Ronneberger2015} blocks (see \cref{sec:denoising-blocks} for more details about these blocks). Multi-scale, U-Net-like architectures have been shown to have the ability to learn misalignment~\cite{Wu2018,Dosovitskiy2015}. Our cascaded architecture increases this capacity of handling movement even further. In contrast to~\cite{Tassano2019}, our architecture is trained end-to-end without optical flow alignment, which avoids distortions and artifacts due to erroneous flow. As a result, we are able to eliminate a costly dedicated motion compensation stage without sacrificing performance. This leads to an important reduction of runtimes: our algorithm runs three orders of magnitude faster than VNLB, and an order of magnitude faster than DVDnet and VNLnet.
	
	\Cref{fig:diagram-overview} displays a diagram of the architecture of our method. When denoising a given frame at time $ t $, $ \tilde{\mathbf{I}}_t $, its $ 2T=4 $ neighboring frames are also taken as inputs. That is, the inputs of the algorithm will be $ \left \{ \tilde{\mathbf{I}}_{t-2},\, \tilde{\mathbf{I}}_{t-1},\, \tilde{\mathbf{I}}_{t},\, \tilde{\mathbf{I}}_{t+1},\, \tilde{\mathbf{I}}_{t+2} \right \} $. The model is composed of different spatio-temporal denoising blocks, assembled in a cascaded two-step architecture. These denoising blocks are all similar, and consist of a  modified {U-Net} model which takes three frames as inputs. The three blocks in the first denoising step share the same weights, which leads to a reduction of memory requirements of the model and facilitates the training of the network. Similar to~\cite{Zhang2017a,Gharbi2016}, a noise map is also included as input, which allows the processing of spatially varying noise~\cite{Tassano2019a}. In particular, the noise map is a separate input which provides information to the network about the distribution of the noise at the input. This information is encoded as the expected per-pixel standard deviation of this noise. For instance, when denoising Gaussian noise, the noise map will be constant; when denoising Poisson noise, the noise map will depend on the intensity of the image. Indeed, the noise map can be used as a user-input parameter to control the trade-off between noise removal vs. detail preservation (see for example the online demo in~\cite{Tassano2019a}). In other cases, such as JPEG denoising, the noise map can be estimated by means of an additional CNN~\cite{Guo2019}. The use of a noise map has been shown to improve denoising performance, particularly when treating spatially variant noise~\cite{Brooks2019}. Contrary to other denoising algorithms, our denoiser takes no other parameters as inputs apart from the image sequence and the estimation of the input noise.
	
	Observe that experiments presented in this paper focus on the case of additive white Gaussian noise (AWGN). Nevertheless, this algorithm can be extended to other types of noise, e.g.\ spatially varying noise (e.g.\ Poissonian). Let $ \mathbf{I} $ be a noiseless image, while $\tilde{\mathbf{I}}$ is its noisy version corrupted by a realization of zero-mean white Gaussian noise $ \mathbf{N} $ of standard deviation $ \sigma $, then
	\begin{equation}\label{eq:noisemod}
	\tilde{\mathbf{I}}=\mathbf{I}+\mathbf{N} \text{ .}
	\end{equation}

	\subsection{Denoising blocks}
	\label{sec:denoising-blocks}	
	Both denoising blocks displayed in \cref{fig:diagram-overview}, \textit{Denoising Block 1} and \textit{Denoising Block 2}, consist of a modified U-Net architecture. All the instances of \textit{Denoising Block 1} share the same weights. U-Nets are essentially a multi-scale encoder-decoder architecture, with skip-connections~\cite{He2016} that forward the output of each one of the encoder layers directly to the input of the corresponding decoder layers. A more detailed diagram of these blocks is shown in \cref{fig:diagram-denoiser}. Our denoising blocks present some differences with respect to the standard U-Net:
	\begin{itemize}
		\item{The encoder has been adapted to take three frames and a noise map as inputs}
		\item{The upsampling in the decoder is performed with a \textit{PixelShuffle} layer~\cite{Shi2016}, which helps reducing gridding artifacts. Please see the supplementary materials for more information about this layer.}
		\item{The merging of the features of the encoder with those of the decoder is done with a pixel-wise addition operation instead of a channel-wise concatenation. This results in a reduction of memory requirements}
		\item{Blocks implement residual learning---with a residual connection between the central noisy input frame and the output---, which has been observed to ease the training process~\cite{Tassano2019a}}
	\end{itemize}
	The design characteristics of the denoising blocks make a good compromise between performance and fast running times. These denoising blocks are composed of a total of $ D = 16 $ convolutional layers. In most layers, the outputs of its convolutional layers are followed by point-wise \textit{ReLU}~\cite{Krizhevsky2012} activation functions $ ReLU(\cdot) = \max (\cdot, 0) $, except for the last layer. Batch normalization layers (\textit{BN}~\cite{Ioffe2015}) are placed between the convolutional and \textit{ReLU} layers.
	
	\section{Discussion}
	\label{sec:discussion}
	%
	Explicit flow estimation is avoided in FastDVDnet. However, in order to maintain performance, we needed to introduce a number of techniques to handle motion and to effectively employ temporal information. These techniques are discussed further in this section. Please see the supplementary materials for more details about ablation studies.
	%
	
	\subsection{Two-step denoising}
	\label{ssec:two-step}
\begin{figure}[!t]
		\centering
		\includegraphics[width=0.9\linewidth]{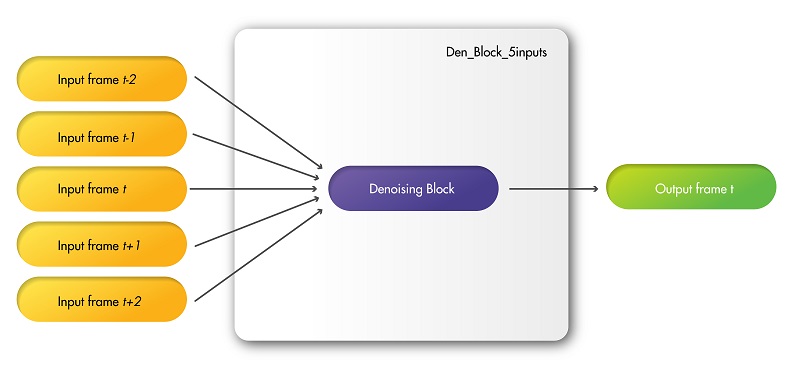}%
		\caption{Architecture of the \textit{Den\_Block\_5inputs} denoiser.}
		\label{fig:architecture-5in}
	\end{figure}

	Similarly to DVDnet and ViDeNN, FastDVDnet features a two-step cascaded architecture. The motivation behind this is to effectively employ the information existent in the temporal neighbors, and to enforce the temporal correlation of the remaining noise in output frames. To prove that the two-step denoising is a necessary feature, we conducted the following experiment: we modified a \textit{Denoising Block} of FastDVDnet (see \cref{fig:diagram-denoiser}) to take five frames as inputs instead of three, which we will refer to as \textit{Den\_Block\_5inputs}. In this way, the same amount of temporal neighboring frames are considered and the same information as in FastDVDnet is processed by this new denoiser. A diagram of the architecture of this model is shown in \cref{fig:architecture-5in}. We then trained this new model and compared the results of denoising of sequences against the results of FastDVDnet (see \cref{sec:training-details} for more details about the training process). 
	
	It was observed that the cascaded architecture of FastDVDnet presents a clear advantage on \textit{Den\_Block\_5inputs}, with differences in PSNR of up to $ 0.9dB $. Please refer to the supplementary materials for more details. Additionally, results by \textit{Den\_Block\_5inputs} present a sharp increase on temporal artifacts---flickering. Despite it being a multi-scale architecture, \textit{Den\_Block\_5inputs} cannot handle the motion of objects in the sequences as well as the two-step architecture of FastDVDnet can. Overall, the two-step architecture shows superior performance with respect to the one-step architecture.
	%
	%
	%
	
	\subsection{Multi-scale architecture and end-to-end training}
	\label{ssec:ms-end-to-end}
	In order to investigate the importance of using multi-scale denoising blocks in our architecture, we conducted the following experiment: we modified the FastDVDnet architecture by replacing its \textit{Denoising Blocks} by the denoising blocks of DVDnet. This results in a two-step cascaded architecture, with single-scale denoising blocks, trained end-to-end, and with no compensation of motion in the scene. In our tests, it was observed that the usage of multi-scale denoising blocks improves denoising results considerably. Please refer to the supplementary materials for more details.
	
	We also experimented with training the multi-scale denoising blocks in each step of FastDVDnet separately---as done in DVDnet. Although the results in this case certainly improved with respect to the case of the single-scale denoising blocks described above, a noticeable flickering remained in the outputs. Switching from this separate training to an end-to-end training helped to reduce temporal artifacts considerably.

	\subsection{Handling of motion}
	\label{ssec:mc}
\begin{figure*}[!t]
		\centering
		\subfloat[]{%
			\includegraphics[width=0.225\linewidth]{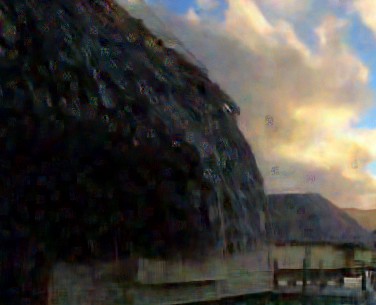}%
			\includegraphics[width=0.225\linewidth]{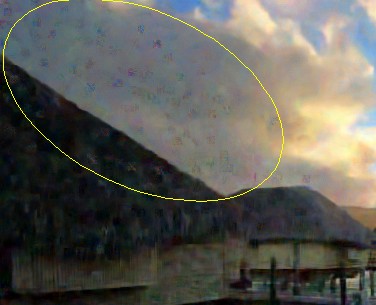}%
			\includegraphics[width=0.225\linewidth]{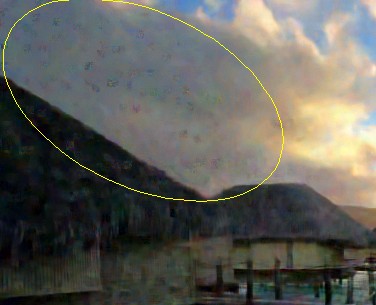}%
			\label{fig:mc-vbm4d}}%
		\hfil
		\subfloat[]{%
			\includegraphics[width=0.225\linewidth]{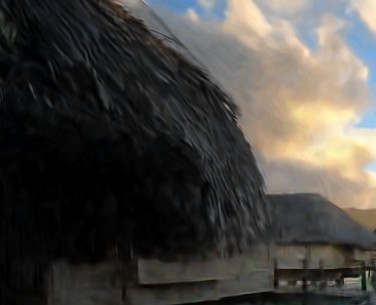}%
			\includegraphics[width=0.225\linewidth]{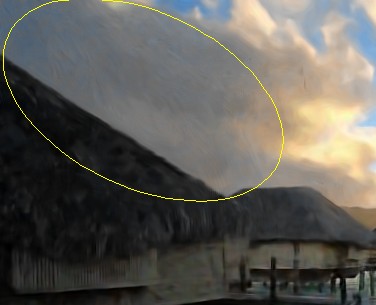}%
			\includegraphics[width=0.225\linewidth]{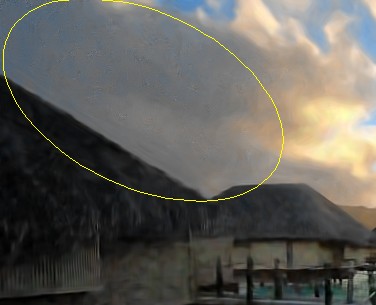}%
			\label{fig:mc-vnlb}}%
		\hfil
		\subfloat[]{%
			\includegraphics[width=0.225\linewidth]{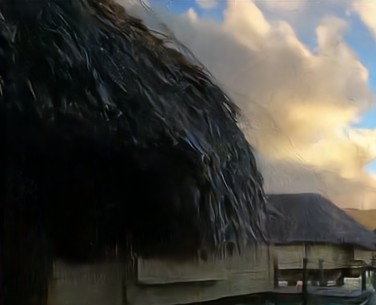}%
			\includegraphics[width=0.225\linewidth]{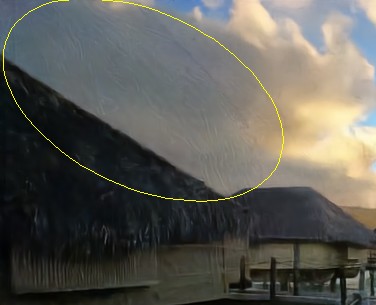}%
			\includegraphics[width=0.225\linewidth]{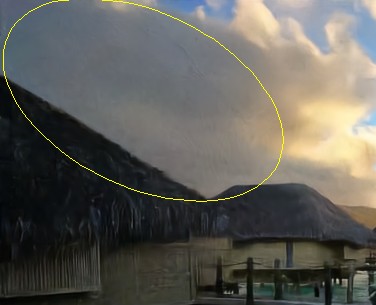}%
			\label{fig:mc-dvdnet}}%
		\hfil
		\subfloat[]{%
			\includegraphics[width=0.225\linewidth]{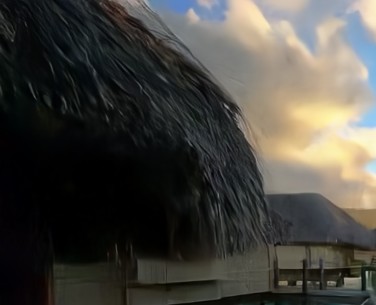}%
			\includegraphics[width=0.225\linewidth]{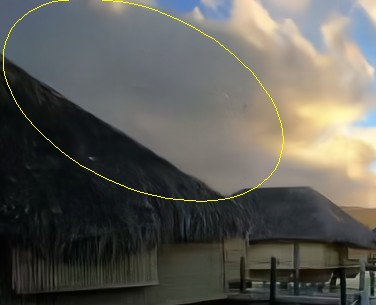}%
			\includegraphics[width=0.225\linewidth]{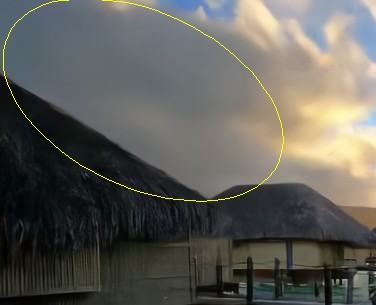}%
			\label{fig:mc-fastdvd}}%
		\caption{\textit{Motion artifacts due to occlusion}. Three consecutive frames of the results of the 'hypersmooth' sequence, $ \sigma=50 $ (a) V-BM4D. (b) VNLB. (c) DVDnet. (d) FastDVDnet. Video denoising algorithms that depend explicitly on motion estimation techniques often present artifacts due to erroneous flow in challenging cases. In the example above, the occlusion of the front building leads to motion artifacts in the results of V-BM4D, VNLB, and DVDnet. Explicit motion compensation is avoided in the architecture of FastDVDnet. Indeed, the network is able to implicitly handle  motion due to its design characteristics. Best viewed in digital format.}
		\label{fig:motion-compensation}
	\end{figure*}

	Apart from the reduction of runtimes, avoiding the use of motion compensation by means of optical flow has an additional benefit. Video denoising algorithms that depend explicitly on motion estimation techniques often present artifacts due to erroneous flow in challenging cases, such as occlusions or strong noise.
	The different techniques discussed in this section---namely a multi-scale of the denoising blocks, the cascaded two-step denoising architecture, and end-to-end training---not only provide FastDVDnet the ability to handle motion, but also help avoid artifacts related to erroneous flow estimation. Also, and similarly to~\cite{Zhang2017,Tassano2019,Tassano2019a}, the denoising blocks of FastDVDnet implement residual learning, which helps to improve the quality of results a step further. \Cref{fig:motion-compensation} shows an example on artifacts due to erroneous flow on three consecutive frames and of how the multi-scale architecture of FastDVDnet is able to avoid them. 
	
	\section{Training details}
	\label{sec:training-details}
	The training dataset consists of input-output pairs $$ P_t^j = \left \{  \left( ( \,S_t^j ,\: \mathbf{M}^j\, ), \, \mathbf{{I}}_t^j \right )\right \}_{j=0}^{m_t} \, ,$$ where $ S_t^j = (\mathbf{\tilde{I}}_{t-2}^j,\:\mathbf{\tilde{I}}_{t-1}^j,\:\mathbf{\tilde{I}}_{t}^j,\:\mathbf{\tilde{I}}_{t+1}^j,\:\mathbf{\tilde{I}}_{t+2}^j) $ is a collection of $ 2T+1=5 $ spatial patches cropped at the same location in contiguous frames, and $ \mathbf{{I}}^j $ is the clean central patch of the sequence. These are generated by adding AWGN of $ \sigma \in [5, \, 50] $ to clean patches of a given sequence, and the corresponding noise map $ \mathbf{M}^j $ is built in this case constant with all its elements equal to $ \sigma $. Spatio-temporal patches are randomly cropped from randomly sampled sequences of the training dataset.
	
	A total of $ m_t = 384000 $ training samples are extracted from the training set of the DAVIS database~\cite{KhoRohrSch_ACCV2018}. The spatial size of the patches is $ 96 \times 96 $, while the temporal size is $ 2T+1 = 5 $. The spatial size of the patches was chosen such that the resulting patch size in the coarser scale of the \textit{Denoising Blocks} is $ 32\times 32 $. The loss function is
	\begin{equation}\label{eq:temp-loss}
	\mathcal{L} (\theta)= \frac{1}{2{m_t}} \sum_{j=1}^{m_t} \left \| \mathbf{\hat{I}}_{t}^j -  \mathbf{{I}}_{t}^j \right \|^2 \, ,
	\end{equation}
	where $ \mathbf{\hat{I}}_{t}^j = \mathcal{F} (( \,S_t^j ,\: \mathbf{M}^j\, );\,\theta) $ is the output of the network, and $ \theta $ is the set of all learnable parameters.
	
	The architecture has been implemented in PyTorch~\cite{Paszke2017}, a popular machine learning library. The ADAM algorithm \cite{Kingma2015} is applied to minimize the loss function, with all its hyper-parameters set to their default values. The number of epochs is set to $ 80 $, and the mini-batch size is $ 96 $. The scheduling of the learning rate is also common to both cases. It starts at $ 1\mathrm{e}{-3} $ for the first $ 50 $ epochs, then changes to $ 1\mathrm{e}{-4} $ for the following $ 10 $ epochs, and finally switches to $ 1\mathrm{e}{-6} $ for the remaining of the training. In other words, a learning rate step decay is used in conjunction with ADAM. The mix of learning rate decay and adaptive rate methods has also been applied to other deep learning projects~\cite{Szegedy2015,Wilson2017}, usually with positive results. Data is augmented by introducing rescaling by different scale factors and random flips. During the first $ 60 $ epochs, the orthogonalization of the convolutional kernels is applied as a means of regularization. It has been observed that initializing the training with orthogonalization may be beneficial to performance~\cite{Zhang2017a,Tassano2019a}.
	
	\section{Results}
	\label{sec:results}
	\begin{figure*}[!t]
		\centering
		\subfloat[]{\includegraphics[width=0.24\linewidth]{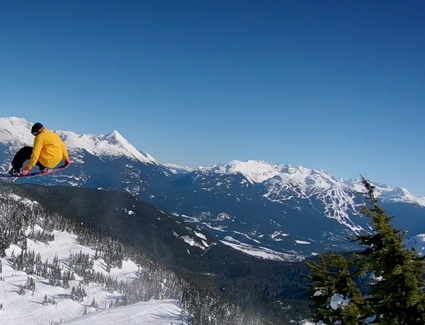}%
			\label{fig:sn-clean}}%
		\hfil
		\subfloat[]{\includegraphics[width=0.24\linewidth]{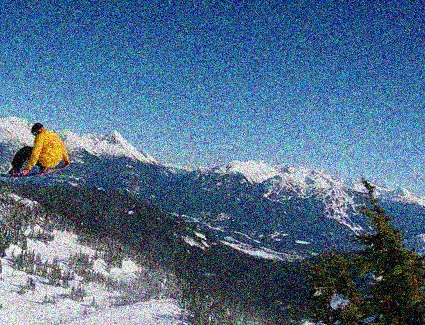}%
			\label{fig:sn-noisy}}%
		\hfil
		\subfloat[]{\includegraphics[width=0.24\linewidth]{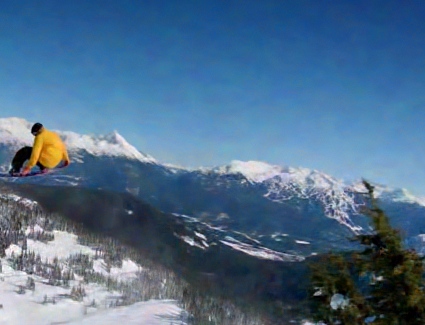}%
			\label{fig:sn-vmb4d}}%
		\hfil	
		\subfloat[]{\includegraphics[width=0.24\linewidth]{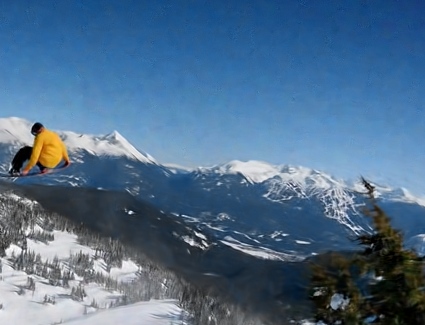}%
			\label{fig:sn-vnlb}}%
		\hfil
		\subfloat[]{\includegraphics[width=0.24\linewidth]{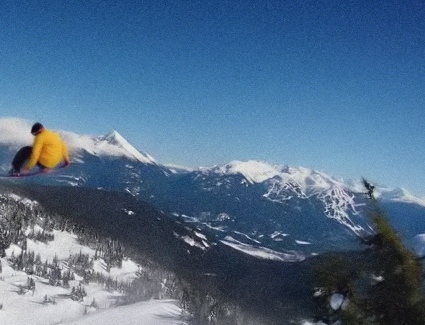}%
			\label{fig:sn-nv}}%
		\hfil
		\subfloat[]{\includegraphics[width=0.24\linewidth]{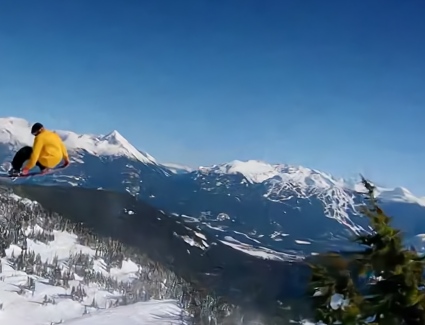}%
			\label{fig:sn-vnlnet}}%
		\hfil
		\subfloat[]{\includegraphics[width=0.24\linewidth]{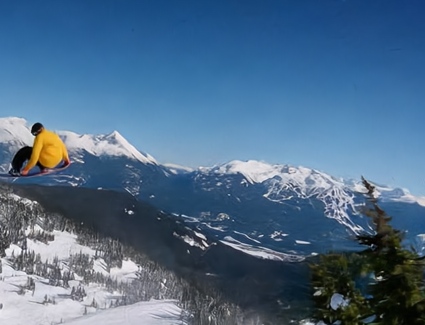}%
			\label{fig:sn-dvdnet}}%
		\hfil
		\subfloat[]{\includegraphics[width=0.24\linewidth]{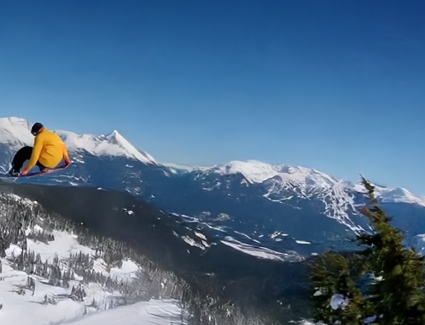}%
			\label{fig:sn-fastdvd}}%
		\hfil
		\caption{\textit{Comparison of results of the 'snowboarding' sequence}. (a) Clean frame. (b) Noisy frame $ \sigma=40 $. (c) V-BM4D. (d) VNLB. (e) NV. (f) VNLnet. (g) DVDnet. (h) FastDVDnet. Patch-based methods (V-BM4D, VNLB, and even VNLnet) struggle with noise in flat areas, such as the sky, and leave behind medium-to-low-frequency noise. This leads to results with noticeable flickering, as the remaining noise is temporally decorrelated. On the other hand, DVDnet and FastDVDnet output very convincing and visually pleasant results. Best viewed in digital format.}
		\label{fig:results-snow}
	\end{figure*}
	\begin{figure*}[!t]
		\centering
		\subfloat[]{\includegraphics[width=0.24\linewidth]{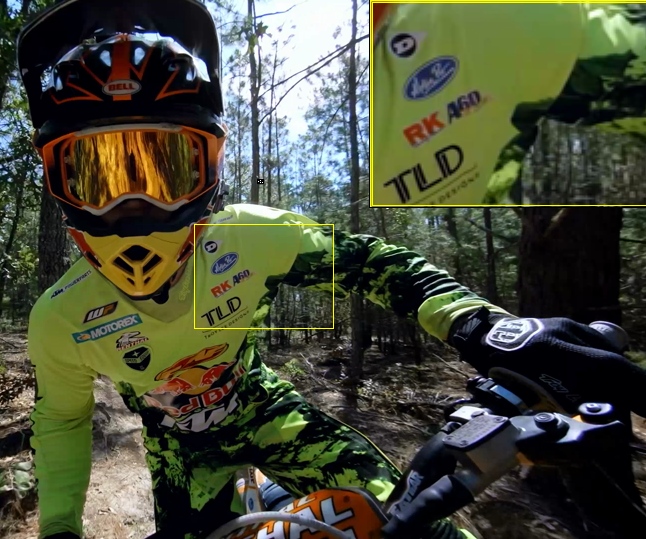}%
			\label{fig:motorbike}}%
		\hfil
		\subfloat[]{\includegraphics[width=0.24\linewidth]{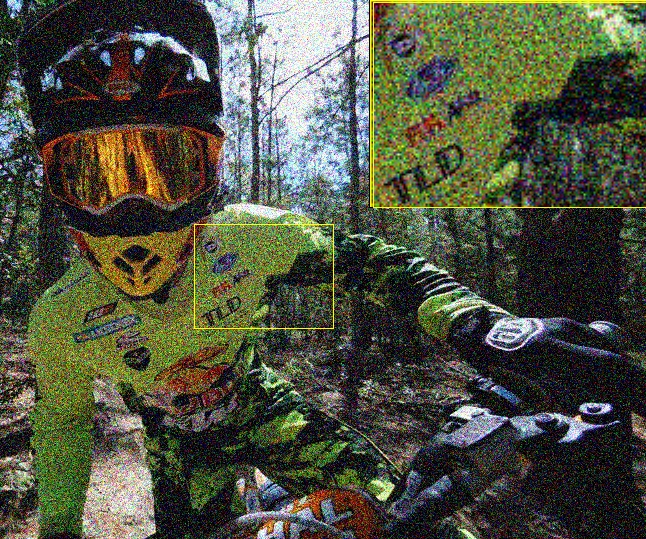}%
			\label{fig:motorbike-noisy}}%
		\hfil
		\subfloat[]{\includegraphics[width=0.24\linewidth]{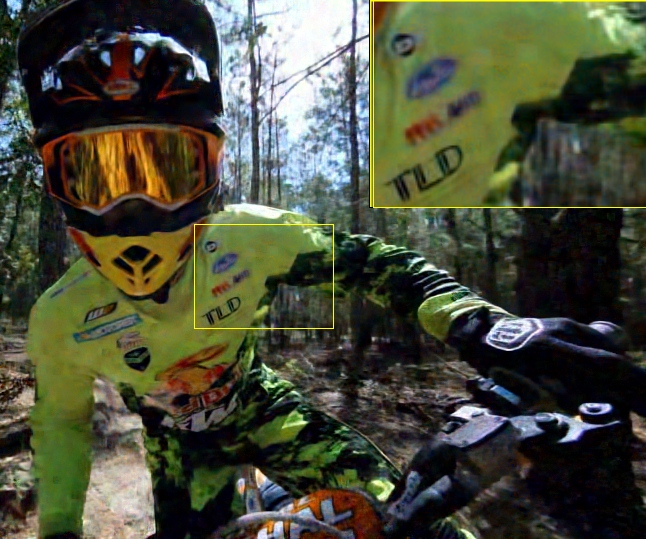}%
			\label{fig:motorbike-vbm4d}}%
		\hfil
		\subfloat[]{\includegraphics[width=0.24\linewidth]{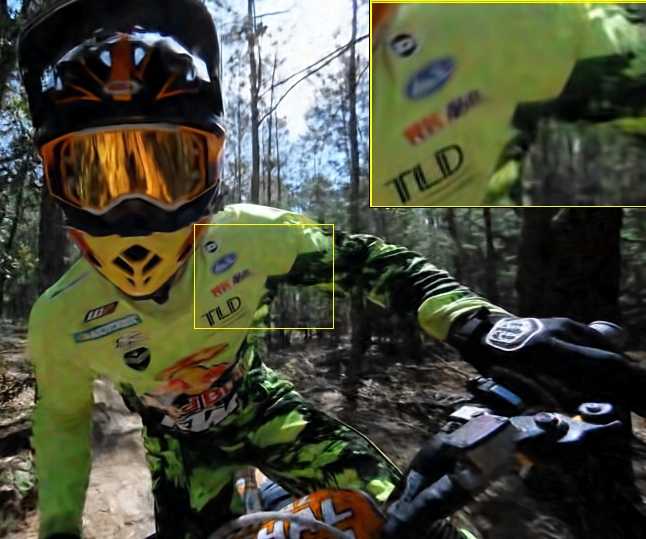}%
			\label{fig:motorbike-vnlb}}%
		\hfil
		\subfloat[]{\includegraphics[width=0.24\linewidth]{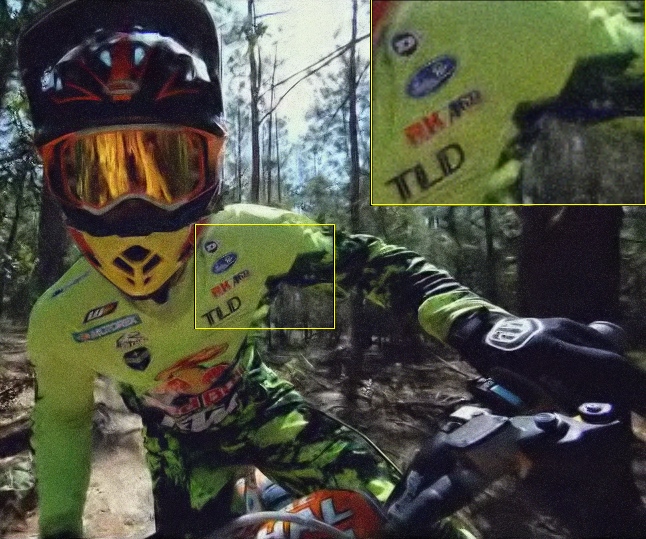}%
			\label{fig:motorbike-nv}}%
		\hfil
		\subfloat[]{\includegraphics[width=0.24\linewidth]{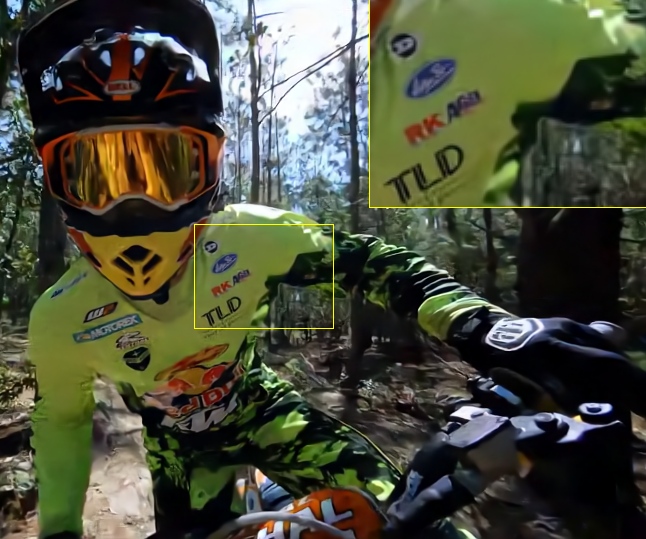}%
			\label{fig:motorbike-vnlnet}}%
		\hfil
		\subfloat[]{\includegraphics[width=0.24\linewidth]{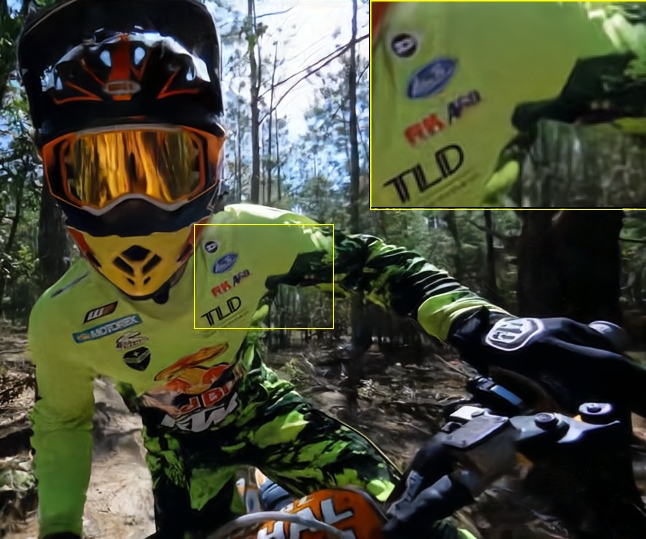}%
			\label{fig:motorbike-dvdnet}}%
		\hfil%
		\subfloat[]{\includegraphics[width=0.24\linewidth]{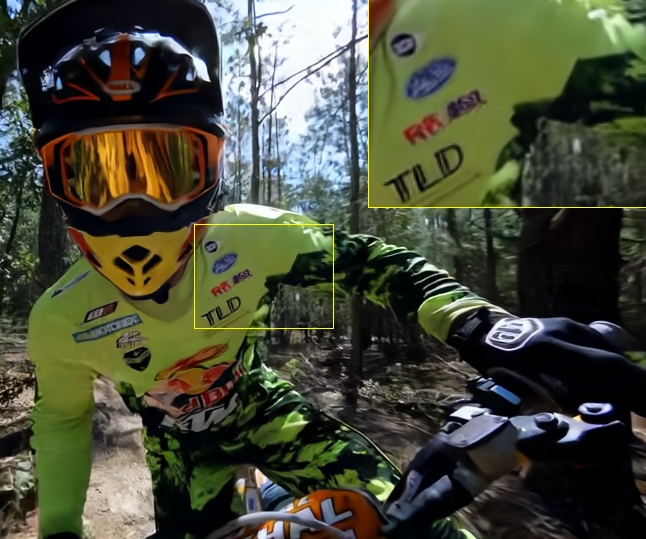}%
			\label{fig:motorbike-fastdvd}}%
		\hfil%
		\caption{\textit{Comparison of results of the 'motorbike' sequence}. (a) Clean frame. (b) Noisy frame $ \sigma=50 $. (c) V-BM4D. (d) VNLB. (e) NV. (f) VNLnet. (g) DVDnet. (h) FastDVDnet. Note the clarity of the denoised text, and the lack of chroma noise for FastDVDnet, DVDnet, and VNLnet. Best viewed in digital format.}
		\label{fig:results-motorbike}
	\end{figure*}

	Two different testsets were used for benchmarking our method: the DAVIS-test testset, and Set8, which is composed of $ 4 $ color sequences from the \textit{Derf’s Test Media collection}\footnote{\url{https://media.xiph.org/video/derf}} and $ 4 $ color sequences captured with a GoPro camera. The DAVIS set contains $ 30 $ color sequences of resolution $ 854 \times 480 $. The sequences of Set8 have been downscaled to a resolution of $ 960 \times 540 $. In all cases, sequences were limited to a maximum of $ 85 $ frames. We used the DeepFlow algorithm~\cite{weinzaepfel:hal-00873592} to compute flow maps for DVDnet and VNLB. VNLnet requires models trained for specific noise levels. As no model is provided for $ \sigma =30 $, no results are shown for this noise level in either of the tables. We also compare our method to a commercial blind denoising software, Neat Video (NV \cite{neatvideo19}). For NV, its automatic noise profiling settings were used to manually denoise the sequences of Set8. Note that  values shown are the average for all sequences in the testset, the PNSR of a sequence is computed as the average of the PSNRs of each frame.
	
	In general, both DVDnet and FastDVDnet output sequences which feature remarkable temporal coherence. Flickering rendered by our methods is notably small, especially in flat areas, where patch-based algorithms often leave behind low-frequency residual noise. An example can be observed in \cref{fig:results-snow} (which is best viewed in digital format). Temporally decorrelated low-frequency noise in flat areas appears as particularly bothersome for the viewer. More video examples can be found in the supplementary materials and on the website of the algorithm. The reader is encouraged to watch these examples to compare the visual quality of the results of our methods.
	
	Patch-based methods are prone to surpassing DVDnet and FastDVDnet in sequences with a large portion of repetitive structures as these methods exploit the non-local similarity prior. On the other hand, our algorithms handle non-repetitive textures very well, see e.g.\ the clarity of the denoised text and vegetation in \cref{fig:results-motorbike}.
	
	\Cref{tbl:results-set8} shows a comparison of PSNR and ST-RRED on the Set8 and DAVIS dataset, respectively. The Spatio-Temporal Reduced Reference Entropic Differences (ST-RRED) is a high performing reduced-reference video quality assessment metric~\cite{Soundararajan2013}. This metric not only takes into account image quality, but also temporal distortions in the video. We computed the ST-RRED scores with the implementation provided by the \textit{scikit-video} library\footnote{http://www.scikit-video.org}. 
	
	It can be observed that for smaller values of noise, VNLB performs better on Set8. Indeed, DVDnet tends to over denoise in some of these cases. FastDVDnet and VNLnet are the best performing algorithms on DAVIS for small sigmas in terms of PSNR and ST-RRED, respectively. However, for larger values of noise DVDnet surpasses VNLB. FastDVDnet performs consistently well in all cases, which is a remarkable feat considering that it runs $ 80 $ times faster than DVDnet, $ 26 $ times faster than VNLnet, and more than $ 4000 $ times faster than VNLB (see \cref{sec:running-times}). Contrary to other denoisers based on CNNs---e.g.\ VNLnet---, our algorithms are able to denoise different noise levels with only one trained model. On top of this, the use of methods involve no hand-tuned parameters, since they only take the image sequence and the estimation of the input noise as inputs.
	\Cref{tbl:results-videnn} displays a comparison with ViDeNN. This algorithm has not actually been trained for AWGN, but for clipped AWGN. Then, a FastDVDnet model to denoise clipped AWGN was trained for this case, which we call \textit{FastDVDnet\_clipped}. It can be observed that the performance of FastDVDnet\_clipped is superior to the performance of ViDeNN by a wide margin.
	\begin{table*}[]
		\centering
		\caption{\label{tbl:results-set8}Comparison of PSNR / ST-RRED on the Set8 and DAVIS testset. For PSNR: larger is better; best results are shown in blue, second best in red. For ST-RRED: smaller is better; best results are shown bold.}
		\begin{tabular}[b]{@{}l c c c c c c@{}}
			\toprule[0.8pt]
			\textbf{Set8}      & VNLB                                       & V-BM4D                & NV                    & VNLnet                            & DVDnet                            & FastDVDnet                     \\ \midrule[0.8pt]
			$ \sigma = 10 $ & $ \color{blue}37.26 $ / $ \textbf{2.86}  $ & $ 36.05 $ / $  3.87 $ & $ 35.67 $ / $  3.42 $ & $ \color{red}37.10 $ / $ 3.43 $   & $ 36.08             $ / $  4.16 $ & $ 36.44 $ / $  3.00 $          \\
			$ \sigma = 20 $ & $ \color{blue}33.72 $ / $ \textbf{6.28}  $ & $ 32.19 $ / $  9.89 $ & $ 31.69 $ / $ 12.48 $ & $ \color{red}33.88 $ / $ 6.88 $   & $ 33.49             $ / $  7.54 $ & $ 33.43 $ / $  6.65 $          \\
			$ \sigma = 30 $ & $ \color{red}31.74  $ / $ \textbf{11.53} $ & $ 30.00 $ / $ 19.58 $ & $ 28.84 $ / $ 33.19 $ & -                                 & $ \color{blue}31.79 $ / $ 12.61 $ & $ 31.68 $ / $ 11.85 $          \\
			$ \sigma = 40 $ & $ 30.39 $  / $ 18.57 $          & $ 28.48 $ / $ 32.82 $ & $ 26.36 $ / $ 47.09 $ & $ \color{blue}30.55 $ / $ 19.71 $ & $ \color{blue}30.55 $ / $ 19.05 $ & $ \color{red}30.46 $ / $ \textbf{18.45} $ \\
			$ \sigma = 50 $ & $ 29.24 $             / $ 27.39 $          & $ 27.33 $ / $ 49.20 $ & $ 25.46 $ / $ 57.44 $ & $ 29.47  $ / $ 29.78 $ & $ \color{blue}29.56 $ / $ 27.97 $ & $ \color{red}29.53 $ / $ \textbf{26.75} $ \\ \bottomrule[0.8pt]
		\end{tabular}
		\begin{tabular}[b]{@{}l c c c c c@{}}
			\toprule[0.8pt]
			\textbf{DAVIS}  & VNLB                              & V-BM4D                & VNLnet                        & DVDnet                                     & FastDVDnet                         \\ \midrule[0.8pt]
			$ \sigma = 10 $ & $ \color{blue}38.85 $ / $  3.22 $ & $ 37.58 $ / $  4.26 $ & $ 35.83 $ / $ \textbf{2.81} $ & $ 38.13             $ / $  4.28 $          & $ \color{red}38.71 $ / $  3.49 $   \\
			$ \sigma = 20 $ & $ 35.68            $ / $  6.77 $  & $ 33.88 $ / $ 11.02 $ & $ 34.49 $ / $ \textbf{6.11} $ & $ \color{red}35.70  $ / $  7.54 $          & $ \color{blue}35.77 $ / $  7.46 $  \\
			$ \sigma = 30 $ & $ 33.73            $ / $ 12.08 $  & $ 31.65 $ / $ 21.91 $ & -                             & $ \color{blue}34.08 $ / $ 12.19 $          & $ \color{red}34.04  $ / $ 13.08 $  \\
			$ \sigma = 40 $ & $ 32.32            $ / $ 19.33 $  & $ 30.05 $ / $ 36.60 $ & $ 32.32 $ / $ 18.63 $         & $ \color{blue}32.86 $ / $ \textbf{18.16} $ & $ \color{red}32.82  $ / $ 20.39 $  \\
			$ \sigma = 50 $ & $ 31.13            $ / $ 28.21 $  & $ 28.80 $ / $ 54.82 $ & $ 31.43 $ / $ 28.67 $         & $ \color{red}31.85 $ / $ \textbf{25.63} $  & $ \color{blue}31.86  $ / $ 28.89 $ \\ \bottomrule[0.8pt]
		\end{tabular}
	\end{table*}
	\begin{table}[]
		\centering
		\caption{\label{tbl:results-videnn}Comparison with ViDeNN for clipped AWGN. See the text for more details. For PSNR: larger is better; best results are shown in bold.}
		\begin{tabular}[b]{@{}lc c@{}}
			\toprule[0.8pt]
			\textbf{DAVIS}  & ViDeNN         & FastDVDnet\_clipped          \\
			&  &                     \\ \midrule[0.8pt]
			$ \sigma = 10 $ & $ 37.13 $      & $ \textbf{38.45} $  \\
			$ \sigma = 30 $ & $ 32.24 $      & $ \textbf{33.52}  $ \\
			$ \sigma = 50 $ & $ 29.77 $      & $ \textbf{31.23}  $ \\ \bottomrule[0.8pt]
		\end{tabular}
	\end{table}
	%
	
	\section{Running times}
	\label{sec:running-times}
\begin{figure}[]
		\centering
		\includegraphics[width=0.9\linewidth]{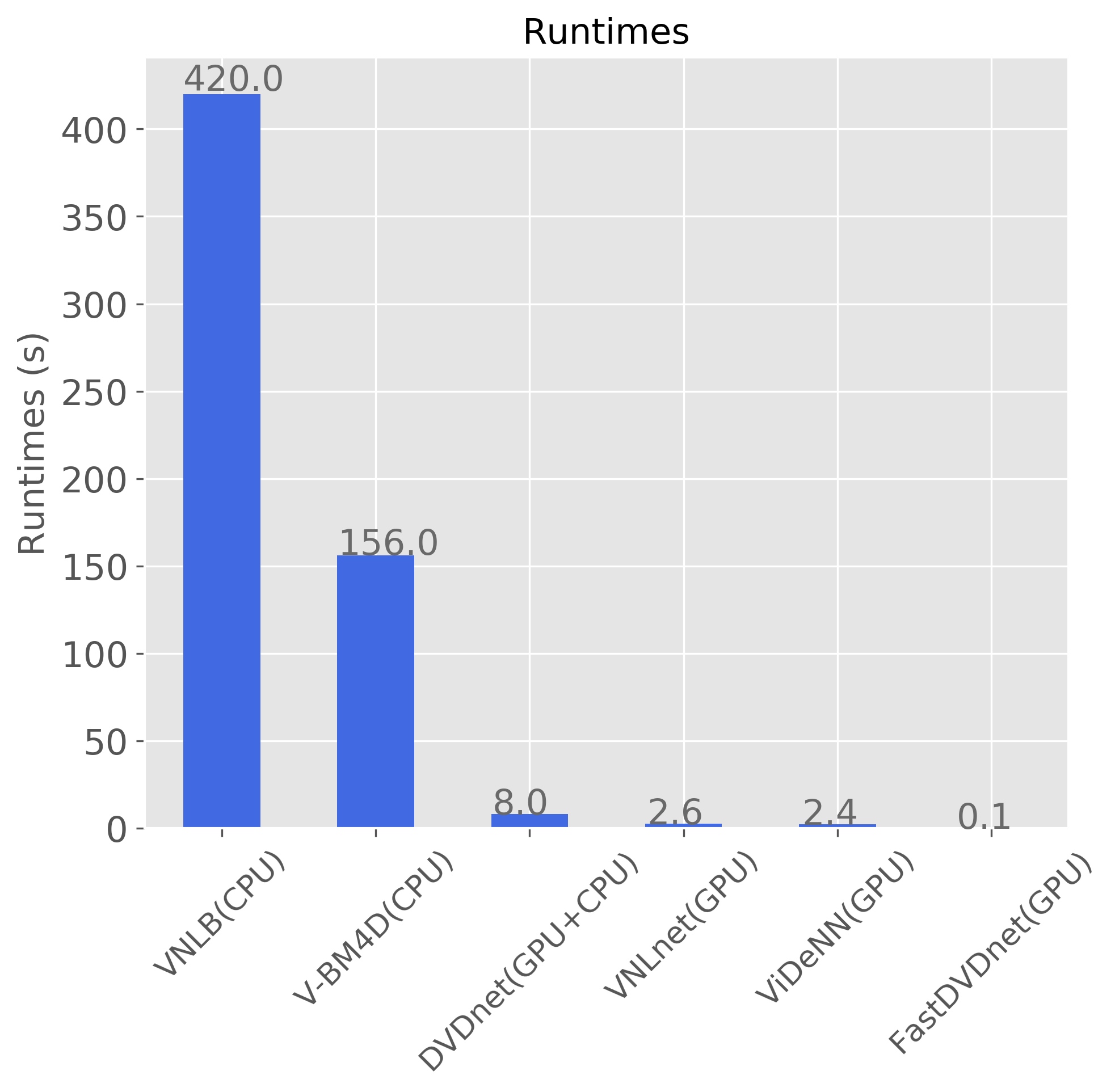}%
		\caption{\label{fig:running-times}\textit{Comparison of running times.} Time to denoise a color frame of resolution $ 960 \times 540 $. Note: values displayed for VNLB do not include the time required to estimate motion.}
	\end{figure}

	Our method achieves fast inference times, thanks to its design characteristics and simple architecture. Our algorithm takes only $ 100ms $ to denoise a $ 960 \times 540 $ color frame, which is more than $ 3 $ orders of magnitude faster than V-BM4D and VNLB, and more than an order of magnitude faster than other CNN algorithms which run on GPU, DVDnet and VNLnet. The algorithms were tested on a server with a Titan Xp NVIDIA GPU card. \Cref{fig:running-times} compares the running times of different state-of-the-art algorithms.
	%
	
	\section{Conclusion}
	\label{sec:conclusions}
	In this paper, we presented FastDVDnet, a state-of-the-art video denoising algorithm. Denoising results of FastDVDnet feature remarkable temporal coherence, very low flickering, and excellent detail preservation. This level of performance is achieved even without a flow estimation step. The algorithm runs between one and three orders of magnitude faster than other state-of-the-art competitors. In this sense, our approach proposes a major step forward towards high quality real-time deep video noise reduction. Although the results presented in this paper hold for Gaussian noise, our method could be extended to denoise other types of noise.
	
	\section*{Acknowledgments}
	Julie Delon would like to thank the support of NVIDIA Corporation for providing us with the Titan Xp GPU used in this research. We thank Anna Murray and José Lezama for their valuable contribution. This work has been partially funded by the French National Research and Technology Agency (ANRT) and GoPro Technology France.
	\clearpage
	{\small
		\bibliographystyle{ieee_fullname}
		\bibliography{refs}
	}	
	
	
	\onecolumn
	\begin{center}
		\textbf{\large Supplemental Materials}
	\end{center}
	\twocolumn
	\setcounter{equation}{0}
	\setcounter{figure}{0}
	\setcounter{table}{0}
	\setcounter{page}{1}
	\setcounter{section}{0}
	\renewcommand{\theequation}{S\arabic{equation}}
	\renewcommand{\thefigure}{S\arabic{figure}}
	
	\section{Two-step denoising}
	\label{ssec:two-step-supl}
\begin{figure}[!t]
		\centering
		\includegraphics[width=\linewidth]{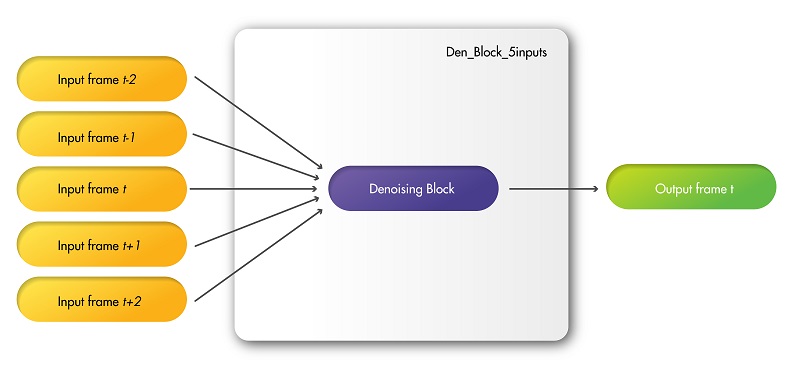}%
		\caption{Architecture of the \textit{Den\_Block\_5inputs} denoiser.}
		\label{fig:architecture-5in-suppl}
	\end{figure}

	FastDVDnet features a two-step cascaded architecture. The motivation behind this is to effectively employ the information existent in the temporal neighbors, and to enforce the temporal correlation of the remaining noise in output frames. To prove that the two-step denoising is a necessary feature, we conducted the following experiment: we modified a \textit{Denoising Block} of FastDVDnet (see the associated paper) to take five frames as inputs instead of three, which we will refer to as \textit{Den\_Block\_5inputs}. In this way, the same amount of temporal neighboring frames are considered and the same information as in FastDVDnet is processed by this new denoiser. A diagram of the architecture of this model is shown in \cref{fig:architecture-5in-suppl}. We then trained this new model and compared the results of denoising of sequences against the results of FastDVDnet. 
	
	\Cref{tbl:results-temp-5in-suppl} displays the PSNRs on four $ 854 \times 480 $ color sequences for both denoisers. It can be observed that the cascaded architecture of FastDVDnet presents a clear advantage on \textit{Den\_Block\_5inputs}, with an average difference of PSNRs of $ 0.95dB $. Additionally, results by \textit{Den\_Block\_5inputs} present a sharp increase on temporal artifacts---flickering. Despite it being a multi-scale architecture, \textit{Den\_Block\_5inputs} cannot handle the motion of objects in the sequences as well as the two-step architecture of FastDVDnet can. Overall, the two-step architecture shows superior performance with respect to the one-step architecture.
	
	\begin{table}[]
		\centering	
		\caption{\label{tbl:results-temp-5in-suppl}Comparison of $ PSNR $ of two denoisers on four sequences. Best results are shown in bold. Note: for this test in particular, neither of these denoisers implement residual learning.}
		\begin{tabular}[b]{@{}l c c c@{}}
			\toprule[0.8pt]
			&             & FastDVDnet     & Den\_Block\_5inputs \\ \midrule[0.8pt]
			$ \sigma = 10 $ & hypersmooth & \textbf{37.34} & 35.64               \\
			& motorbike   & \textbf{34.86} & 34.00               \\
			& rafting     & \textbf{36.20} & 34.61               \\
			& snowboard   & \textbf{36.50} & 34.27               \\ \midrule[0.8pt]
			$ \sigma = 30 $ & hypersmooth & \textbf{32.17} & 31.21               \\
			& motorbike   & \textbf{29.16} & 28.77               \\
			& rafting     & \textbf{30.73} & 30.03               \\
			& snowboard   & \textbf{30.59} & 29.67               \\ \midrule[0.8pt]
			$ \sigma = 50 $ & hypersmooth & \textbf{29.77} & 28.92               \\
			& motorbike   & \textbf{26.51} & 26.19               \\
			& rafting     & \textbf{28.45} & 27.88               \\
			& snowboard   & \textbf{28.08} & 27.37               \\ \bottomrule[0.8pt]
		\end{tabular}
	\end{table}
	
	\section{Multi-scale architecture and end-to-end training}
	\label{ssec:ms-end-to-end-supl}
	In order to investigate the importance of using multi-scale denoising blocks in our architecture, we conducted the following experiment: we modified the FastDVDnet architecture by replacing its \textit{Denoising Blocks} by the denoising blocks of DVDnet. This results in a two-step cascaded architecture, with single-scale denoising blocks, trained end-to-end, and with no compensation of motion in the scene. We will call this new architecture FastDVDnet\_Single. \Cref{tbl:results-fastdvdnet-ffdnet-suppl} shows the PSNRs on four $ 854 \times 480 $ color sequences for both FastDVDnet and FastDVDnet\_Single. It can be seen that the usage of multi-scale denoising blocks improves denoising results considerably. In particular, there is an average difference of PSNRs of $ 0.55dB $ in favor of the multi-scale architecture.
	\begin{table}[]
		\centering
		\caption{\label{tbl:results-fastdvdnet-ffdnet-suppl}Comparison of $ PSNR $ of a single-scale denoiser against a multi-scale denoiser on four sequences. Best results are shown in bold. Note: for this test in particular, neither of these denoisers implement residual learning.}	
		\begin{tabular}[b]{@{}l c c c@{}}
			\toprule[0.8pt]
			&             & FastDVDnet     & FastDVDnet\_Single \\ \midrule[0.8pt]
			$ \sigma = 10 $ & hypersmooth & \textbf{37.34} & 36.61              \\
			& motorbike   & \textbf{34.86} & 34.30              \\
			& rafting     & \textbf{36.20} & 35.54              \\
			& snowboard   & \textbf{36.50} & 35.50              \\ \midrule[0.8pt]
			$ \sigma = 30 $ & hypersmooth & \textbf{32.17} & 31.54              \\
			& motorbike   & \textbf{29.16} & 28.82              \\
			& rafting     & \textbf{30.73} & 30.36              \\
			& snowboard   & \textbf{30.59} & 30.04              \\ \midrule[0.8pt]
			$ \sigma = 50 $ & hypersmooth & \textbf{29.77} & 29.14              \\
			& motorbike   & \textbf{26.51} & 26.22              \\
			& rafting     & \textbf{28.45} & 28.11              \\
			& snowboard   & \textbf{28.08} & 27.56              \\ \bottomrule[0.8pt]
		\end{tabular}
	\end{table}
	
	\section{Ablation studies}
	\label{ssec:ablation}
	A number of modifications with respect to the baseline architecture discussed in the associated paper have been tested, namely:
	\begin{itemize}
		\item the use of \textit{Leaky ReLU}~\cite{Maas2013} or \textit{ELU}~\cite{Clevert2016} instead of \textit{ReLU}. In neither case significant changes in performance were observed, with average differences in PSNR of less than $ 0.05dB $ on all the sequences and standard deviation of noise considered.
		\item optimizing with respect to the Huber loss~\cite{Girshick2015} instead of the $ L_2 $ norm. No significant change of performance was observed. The mean difference in PSNR on all the sequences and standard deviation of noise considered was $ 0.04dB $ in favor of the $ L_2 $ norm case.
		\item removing batch normalization layers. An drop in performance of $ 0.18dB $ on average was observed for this case.
		\item taking more input frames. The baseline model was modified to take 7 and 9 input frames instead of 5. No improvement in performance was observed in neither case. It was also observed an increased difficulty of these models, which have more parameters, to converge during training with respect to the case with 5 input frames.
	\end{itemize}
	
	\section{Upscaling layers}
\begin{figure}[!t]
		\centering
		\includegraphics[width=0.9\linewidth]{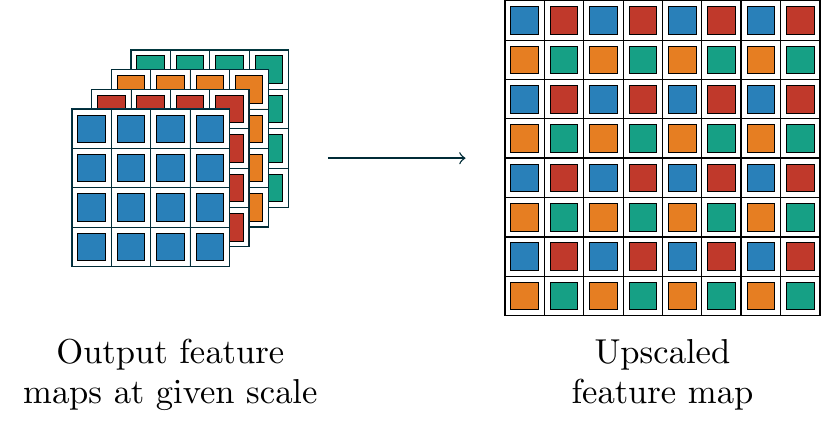}%
		\caption{Upscaling layer.}
		\label{fig:down-up}
	\end{figure}

	In the multi-scale denoising blocks, the upsampling in the decoder is performed with a \textit{PixelShuffle} layer~\cite{Shi2016}. This layer repacks its input of dimension $ 4n_{ch} \times h/2 \times w/2 $ into an output of size $ n_{ch} \times h \times w $, where $ ch,\,h,\,w $ are the number of channels, the height, and the width, respectively. In other words, this layer constructs all the $ 2 \times 2 $ non-overlapping patches of its output with the pixels of different channels of the input, as shown in \cref{fig:down-up}

	\section{Gaussian noise model}
	Recently, a number of algorithms have been proposed for video and burst denoising in low-light conditions, e.g.~\cite{Chen2019,Wang2019,Hasinoff2016}. What is more, some of these works argue that real noise cannot be accurately modeled with a simple Gaussian model. Yet, the algorithm we propose here has been developed for Gaussian denoising because although Gaussian i.i.d.~noise is not utterly realistic, it eases the comparison with other methods on comparable datasets---one of our primary goals. We believe Gaussian denoising is a middle ground where different denoising architectures can be compared fairly. Some networks which are proposed to denoise a specific low-light dataset are designed and overfitted given the image processing pipe of said dataset. In some cases, the comparison against other methods which have not been designed for the given dataset---e.g.~the current version of our method---might not be accurate.
	Nonetheless, low-light denoising is not the main objective of our submission. Rather, it is to show that a simple, yet carefully designed architecture can outperform other more complex methods. We believe that the main challenge to denoising algorithms is the input signal-to-noise ratio. In this regard, the presented results have similar characteristics to low-light videos.
	
	\section{Permutation invariance}
	The algorithm proposed for burst deblurring and denoising in~\cite{Aittala2018} features invariance to the permutation of the ordering of its input frames. One might be tempted to replicate its characteristics in an architecture such as ours to benefit from the advantages of the permutation invariance. However, the application of our algorithm is video denoising---which is not identical to burst denoising. Actually, the order in the input frames is a prior exploited by our algorithm to enforce the temporal coherence in the output sequence. In other words, permutation invariance is not necessarily desirable in our case.
	
	\section{Recursive processing}
	As previously discussed, in practice, the processing of our algorithm is limited to five input frames. Given this limitation, one would wonder if the theoretic performance bound might be lower to that of other solutions based on recursive processing (i.e.~using the output frame in time $ t $ as input to the next frame in time $ t+1 $). Yet, our experience with recursive filtering of videos is that it is difficult for the latter methods to be on par with methods which employ multiple frames as input. Although, in theory, recursive methods are asymptotically more powerful in terms of denoising than multi-frame methods, in practice the performance of recursive methods suffers due to temporal artifacts. Any misalignment or motion compensation artifact which might appear in the output frame at a given time is very likely to appear in all subsequent outputs. An interesting example to illustrate this point is the comparison of the method in~\cite{Ehret2018} versus the video non-local Bayes denoiser (VNLB~\cite{Arias2018}). The former implements a recursive version of VNLB, which results in a lower complexity algorithm, but with very inferior performance with respect to the latter.

\end{document}